\documentclass[final]{cvpr}

\usepackage{times}
\usepackage{epsfig}
\usepackage{graphicx}
\graphicspath{{./fig/}}
\usepackage{amsmath}

\usepackage{amssymb}

\usepackage{algorithm}
\usepackage{algorithmic}

\usepackage{subcaption}

\usepackage[pagebackref=true,breaklinks=true,colorlinks,bookmarks=false]{hyperref}

\def\confYear{CVPR 2021}

\begin{document}

\title{UNIT-DDPM: UNpaired Image Translation \\ with Denoising Diffusion Probabilistic Models}

\author{Hiroshi Sasaki$^1$, Chris G. Willcocks$^1$, Toby P. Breckon$^{1,2}$\\
Department of \{$^1$Computer Science $|$ $^2$Engineering\},
Durham University,
Durham, UK
}

\maketitle

\begin{abstract}
   We propose a novel unpaired image-to-image translation method that uses denoising diffusion probabilistic models without requiring adversarial training.
   Our method, UNpaired Image Translation with Denoising Diffusion Probabilistic Models (UNIT-DDPM), trains a generative model to infer the joint distribution of images over both domains as a Markov chain by minimising a denoising score matching objective conditioned on the other domain. In particular, we update both domain translation models simultaneously, and we generate target domain images by a denoising Markov Chain Monte Carlo approach that is conditioned on the input source domain images, based on Langevin dynamics.
   Our approach provides stable model training for image-to-image translation and generates high-quality image outputs. This enables state-of-the-art Fréchet Inception Distance (FID) performance on several public datasets, including both colour and multispectral imagery, significantly outperforming the  contemporary adversarial image-to-image translation methods.
\end{abstract}

\section{Introduction}

\begin{figure}[tb]
    \begin{center}
      \includegraphics[clip,width=8.5cm]{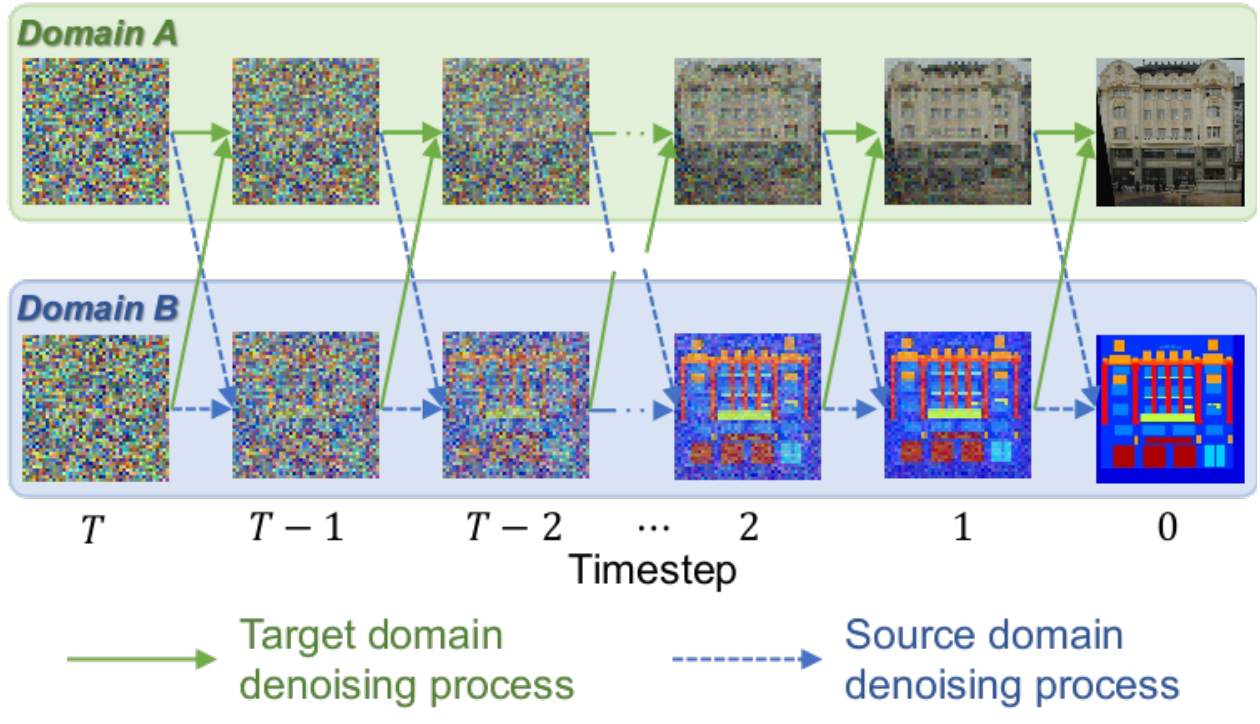}
      \caption{Conceptual illustration of our novel image-to-image translation approach using denoising diffusion probabilistic models.}
      \label{fig:overall}
    \end{center}
  \end{figure}
  
      \begin{table*}[tb]
    \begin{center}
        \caption{Fréchet Inception Distance (FID)~\cite{heusel2017gans} score on different image-to-image translation methods.}
        \begin{tabular}{l r r r r r r r r} \hline
            & \multicolumn{2}{c}{Facades} & \multicolumn{2}{c}{Photos--Maps} & \multicolumn{2}{c}{Summer--Winter} & \multicolumn{2}{c}{RGB--Thermal} \\
            & B$\rightarrow$A & A$\rightarrow$B & B$\rightarrow$A & A$\rightarrow$B & B$\rightarrow$A & A$\rightarrow$B & B$\rightarrow$A & A$\rightarrow$B \\ \hline
            CycleGAN~\cite{zhu2017unpaired} & 232.12 & 265.70 & 216.89 & 150.23 & 121.18 & 133.16 & 338.30 & 169.38\\
            UNIT~\cite{liu2017unit}     & 239.58 & 216.27 & 213.65 & 253.98 & 202.88 & 161.10 & 286.90 & 213.08 \\
            MUNIT~\cite{huang2018multimodal}    & 335.72 & 244.78 & 240.11 & 224.96 & 221.11 & 205.33 & 305.24 & 233.92 \\
            DRIT++~\cite{lee2018diverse}   & 336.76 & 317.23 & 316.42 & 237.94 & 261.82 & 268.45 & 284.21 & 226.35 \\
            UNIT-DDPM (Ours) & \bf{169.95} & \bf{110.13} & \bf{193.06} & \bf{116.23} & \bf{113.70} & \bf{109.98} & \bf{198.85} & \bf{167.70} \\ \hline
        \end{tabular}
        \label{tab:fid}
    \end{center}
\end{table*}
  
Synthesising realistic images is a long-standing goal for computer vision as it enables beneficial and wide-ranging applications such as data augmentation in machine learning tasks, privacy protection, and cost reduction within data acquisition. While there are a wide variety of alternative approaches for image synthesis, such as physical simulation~\cite{gerhart1987thermal}, fractal landscapes~\cite{prusinkiewicz1993fractal}, and image editing~\cite{perez2003poisson}, the use of stochastic generative modelling~\cite{zhu2003statistical} continues to offer remarkable effectiveness in terms of fabricating similar but differing images in specific domains without any domain-specific knowledge. Notably, recent research on the generative modelling focuses on Deep Neural Networks (DNN)~\cite{goodfellow2015deep}, namely deep generative models (DGNN) because of their potential modeling capability of real-world data modalities. Generative Adversarial Networks (GAN)~\cite{goodfellow2014generative}, autoregressive models~\cite{gregor2014deep}, flow-based models such as NICE~\cite{dinh2014nice}, RealNVP~\cite{dinh2016density} and Glow~\cite{kingma2018glow}, Variational Autoencoders (VAE)~\cite{pu2016variational}, and Image Transformer~\cite{parmar2018image} have synthesized remarkably plausible images. Similarly, there have been significant advances in iterative generative models, such as Denoising Diffusion Probabilistic Models (DDPM)~\cite{ho2020denoising} and Noise Conditional Score Networks (NCSN)~\cite{yang2019generative}, which have demonstrated the ability to produce higher quality synthetic images comparable to that of other contemporary approaches, but without having to perform (potentially problematic) adversarial training. To achieve this, many denoising autoencoding models are trained to denoise samples corrupted by various levels of Gaussian noise. Samples are then produced via a Markov Chain Monte Carlo (MCMC) process, starting from white noise, which is progressively denoised and transformed into meaningful high-quality images. This generative Markov Chain process is based on Langevin dynamics~\cite{sohl2015deep} by reversing a forward diffusion process that progressively transforms images into noise.

DGNN has also attracted significant attention within image-to-image (I2I) translation ~\cite{golnaz2017exploring}~\cite{philip2019style}~\cite{isola2017image}~\cite{zhu2017unpaired}. I2I translation is a computer vision task to model the mapping between different visual domains, such as style transfer~\cite{golnaz2017exploring}, colourisation~\cite{dong18colourization}, super-resolution~\cite{ledig2017photo}, photorealistic image synthesis~\cite{chen2017photographic}, and domain adaptation~\cite{murez2018image}. For style transfer, a style transfer network~\cite{golnaz2017exploring} was proposed as the DNN that is trained to transfer the style from one image to another while preserving its semantic content. In addition, style transfer networks are used in the randomisation of image styles~\cite{philip2019style}. For general purposes, Pix2Pix~\cite{isola2017image} employs a GAN to model the mapping function using paired training data. To relax the dependency of the paired training, Cycle-consistent GAN (CycleGAN)~\cite{zhu2017unpaired} was proposed by leveraging cycle consistency to regularise the training. However, such GAN-based approaches require very specific choices in optimisation and architectures to stabilise training and can readily fail to cover all modes of the data distribution~\cite{goodfellow2016nips}.

This paper proposes a new I2I translation approach that uses DDPM as its backend, instead of adversarial networks, in order to mitigate this limitation of unstable training and improve the quality of generated images (Figure~\ref{fig:overall}).
The main contributions of this paper are:
\begin{itemize}
	\item \textit{Dual-domain Markov Chain based Generative Model}~--~a Markov chain I2I translation approach is introduced that approximates the data distribution of both the source and target domains, such that they interrelate with each other (Section~\ref{sec:method}).
	\item \textit{Stable Non-GAN-based Image-to-Image Translation Training}~--~the approach does not require adversarial training, however the model generates realistic outputs that capture high-frequency variations according to the perturbations of various levels of noise (Section~\ref{sec:training}).
	\item \textit{Novel use of Markov Chain Monte Carlo Sampling}~--~the proposed sampling algorithm can be conditioned on unpaired source domain images to synthesise target domain images (Section~\ref{sec:inference}).
	\item \textit{State-Of-The-Art Image-to-Image Translation}~--~the results are found to outperform the prior work of CycleGAN~\cite{zhu2017unpaired}, UNIT~\cite{liu2017unit}, MUNIT~\cite{huang2018multimodal}, and DRIT++~\cite{lee2018diverse} qualitatively and quantitatively for a number of varied benchmark datasets (Facade~\cite{tylevcek2013spatial}, Photos--Maps~\cite{zhu2017unpaired}, Summer--Winter~\cite{zhu2017unpaired}, and RGB--Thermal~\cite{hwang2015multispectral}) (Table~\ref{tab:fid} and Figure~\ref{fig:outputs}), as described in detail in Section~\ref{sec:evaluation}.
\end{itemize}

\section{Related Work}
We review prior work in two relevant topics: image-to-image translation and denoising diffusion probabilistic models.

\subsection{Image-to-Image Translation}
The goal of I2I translation is learning a mapping between images from a source domain and images from a target domain, and I2I translation is generally classified into two types of approach: paired and unpaired.

\subsubsection{Paired Image-to-Image Translation}
Supervised I2I approaches aim to learn the mapping between an input image and an output image using a training set of aligned image pairs.
Earlier work proposed to use a pre-trained CNN and Gram matrices to obtain the perceptual decomposition of images~\cite{gatys2015neural}. This separates the image content and style, enabling style variation whilst preserving the semantic content. Many recent I2I approaches are trained adversarially with a GAN~\cite{goodfellow2014generative}, which is a generative model designed to have a generator and a discriminator component that compete against each other. The generator is trained to map randomised values to real data examples by the discriminator output. The discriminator is simultaneously trained to discriminate real and fake data examples produced by the generator.
Pix2Pix~\cite{isola2017image} provides a general-purposed adversarial framework to transform an image from one domain to another. Instead of an autoencoder, U-Net~\cite{ronneberger2015u} is utilized to share the low-level information between the input and output.
BicycleGAN~\cite{zhu2017toward} combines conditional VAE-GAN (CVAE-GAN) with an approach to recover the latent codes, which improves performance, where the CVAE-GAN reconstructs category specific images~\cite{bao2017cvae}.

\subsubsection{Unpaired Image-to-Image Translation}
While paired I2I translation requires aligned image pairs of source and target domains, unpaired approaches learn source and target image sets which are completely independent without paired examples between the two domains.
CycleGAN~\cite{zhu2017unpaired} is an unpaired I2I translation approach using a GAN. CycleGAN modifies the generator $G$ and discriminator $D$ to transfer from source images $x_s \in X_s$ to target images $x_t \in X_t$. This learns not only a lateral transform $G$, but also bilateral transform paths $G_t(x_s),G_s(x_t)$. In addition, this adopts a new loss measure named a cycle-consistency loss $L_{\text{cyc}}(G_s,G_t)$:
\begin{eqnarray}
 \mathcal{L}_{\text{cyc}}(G_s,G_t) = \mathbb{E}_{x_s \in X_s}[\|G_s(G_t(x_s))-x_s\|_1] \nonumber \\
  + \mathbb{E}_{x_t \in X_t}[\|G_t(G_s(x_t))-x_t\|_1],
  \label{eqn:cycle_consistency}
\end{eqnarray}
which enforces consistency between real images from each domain and their generated counterparts.

Unsupervised Image-to-Image Translation Networks (UNIT)~\cite{liu2017unit} further make a shared-latent space assumption in their method. To address the multimodal problem, Multimodal UNIT (MUNIT)~\cite{huang2018multimodal} and Diverse Image-to-Image Translation via Disentangled Representations (DRIT++)~\cite{lee2018diverse} adopt a disentangled representation, which separates domain-specific attributes and shared content information within images, to further realise diverse I2I translation from unpaired image samples.

\subsection{Denoising Diffusion Probabilistic Models}
Denoising Diffusion Probabilistic Models (DDPM)~\cite{ho2020denoising} sequentially corrupt images with increasing noise, and learn to reverse the corruption as a generative model. In particular, the generative process is defined as the reverse of a Markovian diffusion process which, starting from white noise, progressively denoises the sample into an image.

DDPM models data as latent variable of the form $p_\theta(\text{\bf{x}}_0):=\int p_\theta(\text{\bf{x}}_{0:T})dx_{1:T}$, where $\text{\bf{x}}_0 \sim q(\text{\bf{x}}_0)$ are images, $T$ is the length of the Markov chain, and $\text{\bf{x}}_1,...,\text{\bf{x}}_T$ are latents of the same dimensionally as the images. $p_\theta(\text{\bf{x}}_{0:T})$ is a Markov-chain with learnt Gaussian transitions (the reverse process) where:
\begin{eqnarray}
    p_\theta(\text{\bf{x}}_{0:T}) &:=& p(\text{\bf{x}}_T) \prod\limits_{t=1}^T p_\theta (\text{\bf{x}}_{t-1} | \text{\bf{x}}_t) \label{eqn:rev_proc} \\
    p_\theta (\text{\bf{x}}_{t-1} | \text{\bf{x}}_t) &=& \mathcal{N}(\text{\bf{x}}_{t-1}; \mu_\theta (\text{\bf{x}}_t,t), \Sigma_\theta (\text{\bf{x}}_t,t)) \label{eqn:sig}\\
    p(\text{\bf{x}}_T) &=& \mathcal{N}(\text{\bf{x}}_T; \text{\bf{0}}, \text{\bf{I}} )
\end{eqnarray}
DDPM additionally approximates the posterior $q(\text{\bf{x}}_{1:T} | \text{\bf{x}}_0)$ in the forward process. This Markov-chain gradually adds progressive Gaussian noise to the images:
\begin{eqnarray}
    q(\text{\bf{x}}_{1:T} | \text{\bf{x}}_0) &:=& \prod\limits_{t=1}^T q(\text{\bf{x}}_t | \text{\bf{x}}_{t-1}) \label{eqn:fwd_proc} \\
    q(\text{\bf{x}}_t | \text{\bf{x}}_{t-1}) &=& \mathcal{N}(\text{\bf{x}}_t; \sqrt{\alpha_t}\text{\bf{x}}_{t-1}, (1-\alpha_t)\text{\bf{I}}), \label{eqn:fwd_lerp}
\end{eqnarray}
where $\alpha_t \in \{\alpha_1,...,\alpha_T\}$ are scheduled weights of the noise, therefore Equation~(\ref{eqn:fwd_proc}) gradually adds Gaussian noise according to a variance schedule $\alpha_t$. Equation~(\ref{eqn:fwd_lerp}) is a linear interpolation function of noise and images, which admits sampling $\text{\bf{x}}_t$ at an arbitrary timestep $t$ as:
\begin{eqnarray}
    \text{\bf{x}}_t(\text{\bf{x}}_0, \boldsymbol\epsilon) = \sqrt{\bar{\alpha}_t}\text{\bf{x}}_0+\sqrt{1-\bar{\alpha}_t}\bf{\boldsymbol\epsilon},
\end{eqnarray}
where $\bar{\alpha}_t := \prod\limits_{s=1}^t \alpha_s$ and $\boldsymbol\epsilon \sim \mathcal{N}(\text{\bf{0}}, \text{\bf{I}} )$.
To approximate $p_\theta (\text{\bf{x}}_{t-1} | \text{\bf{x}}_t)$, DDPM optimises the model parameter $\theta$ via denoising score matching (DSM)~\cite{vincent2011connection}. The loss function is thus redefined as a simpler form as:
\begin{eqnarray}
    \mathcal{L}_{\text{simple}}(\theta) {=} \mathbb{E}_{t,\text{\bf{x}}_0,\boldsymbol\epsilon}[\| \boldsymbol\epsilon-\epsilon_\theta(\text{\bf{x}}_t(\text{\bf{x}}_0, \boldsymbol\epsilon),t) \|^2],
    \label{eqn:ddpm_loss}
\end{eqnarray}
where $\epsilon_\theta$ is a non-linear function predicting the added noise $\boldsymbol\epsilon$ from $\text{\bf{x}}_t$ and $t$.
Using an approximated $\epsilon_\theta$, $\mu_\theta$ can be predicted as:
\begin{eqnarray}
    \mu_\theta(\text{\bf{x}}_t,t)=\frac{1}{\sqrt{\alpha}}\left(\text{\bf{x}}_t-\frac{1-\alpha_t}{\sqrt{1-\bar{\alpha}_t}}\epsilon_\theta(\text{\bf{x}}_t,t)\right).
\end{eqnarray}
$\Sigma_\theta$ in Equation (\ref{eqn:sig}) is set as $\Sigma_\theta(\text{\bf{x}}_t,t) = (1-\alpha_t)\text{\bf{I}}$. This admits sampling $\text{\bf{x}}_{t-1}$ from $\text{\bf{x}}_t$:
\begin{eqnarray}
    \text{\bf{x}}_{t-1} = \mu_\theta(\text{\bf{x}}_t,t) + \Sigma_\theta(\text{\bf{x}}_t,t)\boldsymbol\epsilon,
    \label{eqn:ddpm_sampling}
\end{eqnarray}
which leads to being able to sample $\text{\bf{x}}_0$. 

Our method applies the latent information approximated via DDPM to learn different domains of images and hence it connects between the latents of those domains. As a result, it allows gradual samples from noise, progressively denoised to images, within the target domain in such a way that is related to the input source domain images.  
\section{Methodology}
\label{sec:method}
Our aim is to develop I2I translation between different domains of images whose distributions are formed as the joint probability of Equation~(\ref{eqn:rev_proc}) respectively. The method needs to learn the parameters of the models from a given dataset of source and target domains via empirical risk minimisation and subsequently be able to infer the target domain images from the corresponding source domain images. 

\begin{figure}[tb]
    \centering
      \begin{tabular}{c}
        \begin{minipage}{1.0\hsize}
          \includegraphics[clip,width=8.5cm]{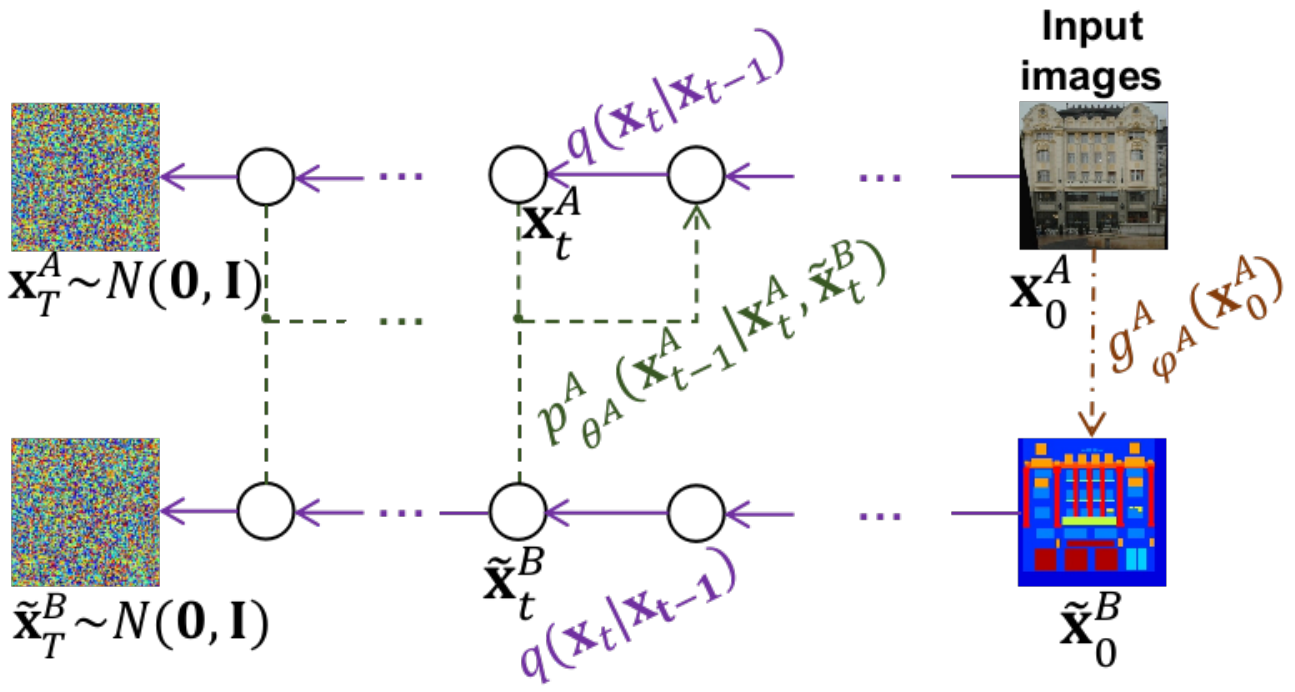}
          \subcaption{Model Training}
        \end{minipage} \\
        \begin{minipage}{0.02\hsize}
          \vspace{1mm}
        \end{minipage} \\
        \begin{minipage}{1.0\hsize}
          \includegraphics[clip,width=8.0cm]{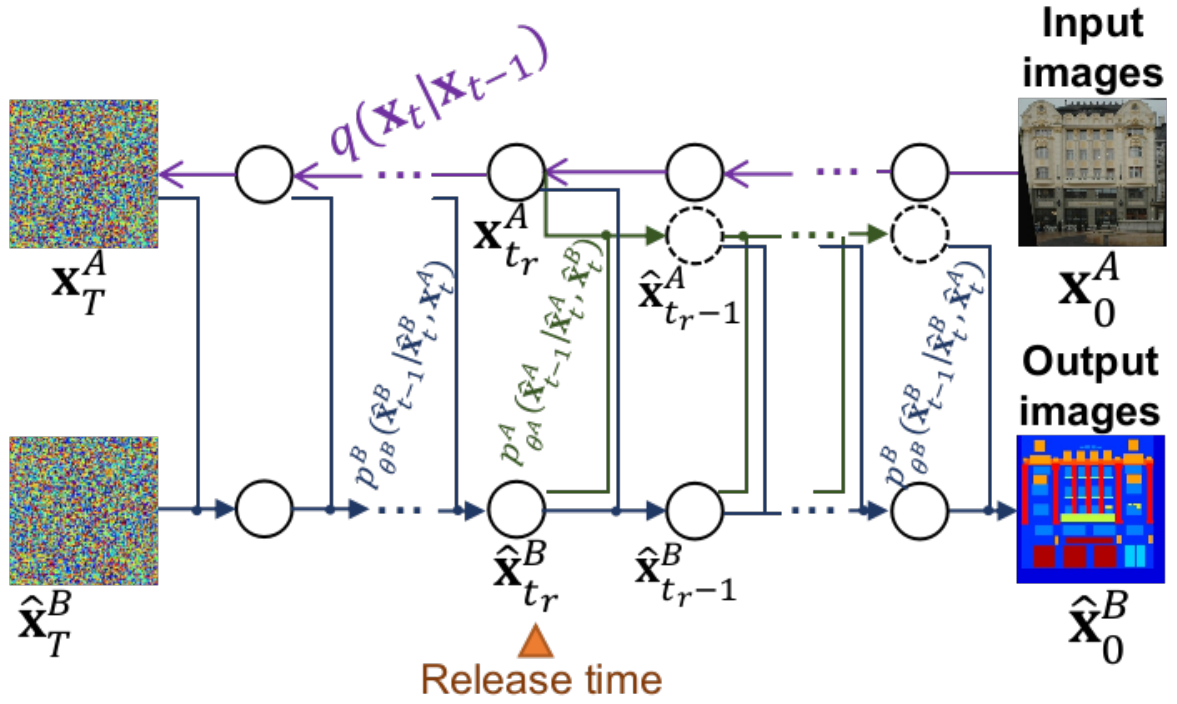}
          \subcaption{Image Translation (Inference)}
        \end{minipage} \\
      \end{tabular}
      \caption{The processing flow of our method. Model training (top) and inference of image translation (inference) (bottom).}
      \label{fig:flow}
  \end{figure}
  
  \begin{algorithm}[tb]
    \caption{UNIT-DDPM Training}
    \begin{algorithmic}[1]
        \REPEAT
            \STATE $\text{\bf{x}}^A_0 \in \mathcal{X}^A, \text{\bf{x}}^B_0 \in \mathcal{X}^B$
            \STATE $\tilde{\text{\bf{x}}}^A_0 \leftarrow g^B_{\phi^B}(\text{\bf{x}}^B_0), \tilde{\text{\bf{x}}}^B_0 \leftarrow g^A_{\phi^A}(\text{\bf{x}}^A_0)$
            \STATE $t^A, t^B \sim \text{Uniform}(\{1,...,T\})$
            \STATE $\boldsymbol\epsilon^A, \boldsymbol\epsilon^B \sim \mathcal{N}(\text{0},\text{I})$
            \STATE $\text{\bf{x}}^A_{t^A} \leftarrow \sqrt{\bar{\alpha}_{t^A}}\text{\bf{x}}^A_0+\sqrt{1-\bar{\alpha}_{t^A}}\boldsymbol\epsilon^A$ \\
                $\text{\bf{x}}^B_{t^B} \leftarrow \sqrt{\bar{\alpha}_{t^B}}\text{\bf{x}}^B_0+\sqrt{1-\bar{\alpha}_{t^B}}\boldsymbol\epsilon^B$
            \STATE $\tilde{\text{\bf{x}}}^A_{t^B} \leftarrow \sqrt{\bar{\alpha}_{t^B}}\tilde{\text{\bf{x}}}^A_0+\sqrt{1-\bar{\alpha}_{t^B}}\boldsymbol\epsilon^A$ \\
                $\tilde{\text{\bf{x}}}^B_{t^A} \leftarrow \sqrt{\bar{\alpha}_{t^A}}\tilde{\text{\bf{x}}}^B_0+\sqrt{1-\bar{\alpha}_{t^A}}\boldsymbol\epsilon^B$
            \STATE \text{Take gradient descent step on} \\
            $\nabla_{\theta^A, \theta^B}[ \| \boldsymbol\epsilon^A {-} \epsilon^A_{\theta^A}(\text{\bf{x}}^A_{t^A}, \tilde{\text{\bf{x}}}^B_{t^A}, t^A) \|^2$ \\
            ~~~~~~~~~~~~$+ \| \boldsymbol\epsilon^B{-}\epsilon^B_{\theta^B}(\text{\bf{x}}^B_{t^B}, \tilde{\text{\bf{x}}}^A_{t^B}, t^B) \|^2 ]$
            \STATE \text{Take gradient descent step on} \\
            $\nabla_{\phi^A, \phi^B}[ \| \boldsymbol\epsilon^A {-} \epsilon^A_{\theta^A}(\text{\bf{x}}^A_{t^A}, \tilde{\text{\bf{x}}}^B_{t^A}, t^A) \|^2$ \\
            ~~~~~~~~~~~~$+\| \boldsymbol\epsilon^A {-} \epsilon^A_{\theta^A}(\tilde{\text{\bf{x}}}^A_{t^B}, \text{\bf{x}}^B_{t^B}, t^B) \|^2$ \\
            ~~~~~~~~~~~~$+\| \boldsymbol\epsilon^B {-} \epsilon^B_{\theta^B}(\text{\bf{x}}^B_{t^B}, \tilde{\text{\bf{x}}}^A_{t^B}, t^B) \|^2$ \\
            ~~~~~~~~~~~~$+\| \boldsymbol\epsilon^B {-} \epsilon^B_{\theta^B}(\tilde{\text{\bf{x}}}^B_{t^A}, \text{\bf{x}}^A_{t^A}, t^A) \|^2$ \\
            ~~~~~~~~~~~~$+\lambda_{\text{cyc}}\| g^B_{\phi^B}(\tilde{\text{\bf{x}}}^B_0) - \text{\bf{x}}^A_0 \|^2$ \\
            ~~~~~~~~~~~~$+\lambda_{\text{cyc}}\| g^A_{\phi^A}(\tilde{\text{\bf{x}}}^A_0) - \text{\bf{x}}^B_0 \|^2]$
        \UNTIL converged
    \end{algorithmic}
    \label{alg:training}
\end{algorithm}

\subsection{Model Training}
\label{sec:training}
Assuming a source domain  $\text{\bf{x}}^A_0 \in \mathcal{X}^A$ and a target domain $\text{\bf{x}}^B_0 \in \mathcal{X}^B$, we iteratively optimise the reverse process in each domain $p^A_{\theta^A}, p^B_{\theta^B}$ and the domain translation functions $\tilde{\text{\bf{x}}}^B_{0} = g^A_{\phi^A}(\text{\bf{x}}^A_{0}), \tilde{\text{\bf{x}}}^A_{0} = g^B_{\phi^B}(\text{\bf{x}}^B_{0})$, which are only used in the model training to transfer the domain A to B and B to A respectively, via DSM (Figure~\ref{fig:flow} (top)).
To enable translation between source domain and target domain image pairs, $p^A_{\theta^A}, p^B_{\theta^B}$ is modified as $p^A_{\theta^A}(\text{\bf{x}}^A_{t-1} | \text{\bf{x}}^A_{t}, \tilde{\text{\bf{x}}}^B_{t}), p^B_{\theta^B}(\text{\bf{x}}^B_{t-1} | \text{\bf{x}}^B_{t}, \tilde{\text{\bf{x}}}^A_{t})$ such as to be conditional on the generated images.
On the reverse process optimisation step, the model parameters $\theta^A, \theta^B$ are updated to minimise the loss function based on Equation~(\ref{eqn:ddpm_loss}), which is rewritten as:
\begin{eqnarray}
    && \mathcal{L}_{\theta}(\theta^A,\theta^B) = \nonumber \\
    && \qquad \mathbb{E}_{t,\text{\bf{x}}^A_0,\boldsymbol\epsilon}[\| \boldsymbol\epsilon-\epsilon^A_{\theta^A}(\text{\bf{x}}_t(\text{\bf{x}}^A_0, \boldsymbol\epsilon), \tilde{\text{\bf{x}}}^B_{t},t) \|^2] \nonumber \\
    && \qquad {+} \mathbb{E}_{t,\text{\bf{x}}^B_0,\boldsymbol\epsilon}[\| \boldsymbol\epsilon{-}\epsilon^B_{\theta^B}(\text{\bf{x}}_t(\text{\bf{x}}^B_0, \boldsymbol\epsilon), \tilde{\text{\bf{x}}}^A_{t},t) \|^2]
    \label{eqn:train_denoising}
\end{eqnarray}
The parameters of the domain translation functions $\phi^A, \phi^B$ are updated to minimise the DSM objective fixing $\theta^A, \theta^B$ where:
\begin{eqnarray}
    && \mathcal{L}_{\epsilon^\phi}(\phi^A,\phi^B) = \nonumber \\
    && \mathbb{E}_{t,\text{\bf{x}}^B_0,\boldsymbol\epsilon}[\| \boldsymbol\epsilon-\epsilon^A_{\theta^A}(\text{\bf{x}}_t(g^B_{\phi^B}(\text{\bf{x}}^B_0), \boldsymbol\epsilon), \text{\bf{x}}_t(\text{\bf{x}}^B_0, \boldsymbol\epsilon), t) \|^2 \nonumber \\
    && \qquad + \| \boldsymbol\epsilon-\epsilon^B_{\theta^B}(\text{\bf{x}}_t(\text{\bf{x}}^B_0, \boldsymbol\epsilon), g^B_{\phi^B}(\text{\bf{x}}_t(\text{\bf{x}}^B_0), \boldsymbol\epsilon), t) \|^2 ] \nonumber \\
    && + \mathbb{E}_{t,\text{\bf{x}}^A_0,\boldsymbol\epsilon}[\| \boldsymbol\epsilon-\epsilon^B_{\theta^B}(\text{\bf{x}}_t(g^A_{\phi^A}(\text{\bf{x}}^A_0), \boldsymbol\epsilon), \text{\bf{x}}_t(\text{\bf{x}}^A_0, \boldsymbol\epsilon), t) \|^2 \nonumber \\
    && \qquad + \| \boldsymbol\epsilon-\epsilon^A_{\theta^A}(\text{\bf{x}}_t(\text{\bf{x}}^A_0, \boldsymbol\epsilon), g^A_{\phi^A}(\text{\bf{x}}_t(\text{\bf{x}}^A_0), \boldsymbol\epsilon), t) \|^2 ]
    \label{eqn:train_denoising_trans}
\end{eqnarray}
 In addition, the training is regularised by the cycle-consistency loss that is proposed in~\cite{zhu2017unpaired} to make both domain translation models bijective. The cycle-consistency loss, Equation~(\ref{eqn:cycle_consistency}) is rewritten as:
 \begin{eqnarray}
    \mathcal{L}_{\text{cyc}^\phi}(\phi^A,\phi^B)= \mathbb{E}_{\text{\bf{x}}^B_0}[ \| g^A_{\phi^A}(g^B_{\phi^B}(\text{\bf{x}}^B_0))-\text{\bf{x}}^B_0 \|_1 ] \nonumber \\
    \qquad +\mathbb{E}_{\text{\bf{x}}^A_0}[\| g^B_{\phi^B}(g^A_{\phi^A}(\text{\bf{x}}^A_0))-\text{\bf{x}}^A_0 \|_1 ]
    \label{eqn:train_cyc}
\end{eqnarray}
 The loss function is thus described as follows:
\begin{eqnarray}
    \mathcal{L}_{\phi}(\phi^A,\phi^B) = \mathcal{L}_{\epsilon^\phi}(\phi^A,\phi^B) + \lambda_{\text{cyc}}\mathcal{L}_{\text{cyc}^\phi}(\phi^A,\phi^B),
    \label{eqn:train_translation}
\end{eqnarray}
where $\lambda_\text{cyc}$ is a cycle-consistency loss weight.
The overall training process is presented in Algorithm~\ref{alg:training}.

\subsection{Inference of Image Translation}
\label{sec:inference}
Using the trained $\theta^A, \theta^B$, the input images are translated from source to target domain. The domain translation functions are no longer used in inference. Instead, the target domain images are progressively synthesised from Gaussian noise and the noisy source domain images. During sampling, the generative process is conditioned on the input source domain images that are perturbed by the forward process from $t=T$ until an arbitrary timtestep $t_r\in[1,T]$. This is then re-generated by the reverse process from this timestep, which we denote as the \textit{release time} (Figure~\ref{fig:flow} (bottom)). The case of transferring from domain A $\text{\bf{x}}^A_0$ to domain B $\hat{\text{\bf{x}}}^B_0$ is described as:
\begin{eqnarray}
    \hat{\text{\bf{x}}}^B_{t-1} &=& \mu_{\theta^B}(\hat{\text{\bf{x}}}^B_t, \hat{\text{\bf{x}}}^A_t, t) {+} \Sigma_{\theta^B}(\text{\bf{x}}_t,t)\epsilon^B \label{eqn:sample_translation} \\
    \hat{\text{\bf{x}}}^A_{t-1} &=& \left\{ \begin{array}{ll}
        \sqrt{\bar{\alpha}_{t^A}}\text{\bf{x}}^A_0+\sqrt{1-\bar{\alpha}_{t^A}}\boldsymbol\epsilon^A & (t {>} t_r) \\
        \mu_{\theta^A}(\hat{\text{\bf{x}}}^A_t{,}\hat{\text{\bf{x}}}^B_t{,} t) {+} \Sigma_{\theta^A}(\text{\bf{x}}^A_t{,}t)\boldsymbol\epsilon^B & (t {\leq} t_r) \\
    \end{array} \right. \\
    && \hat{\text{\bf{x}}}^B_T, \boldsymbol\epsilon^A, \boldsymbol\epsilon^B \sim \mathcal{N}(\text{0},\text{I})
\end{eqnarray}
The overall translation (inference) process is presented in Algorithm~\ref{alg:translation}.
\begin{algorithm}[tb]
    \caption{UNIT-DDPM Inference ($\mathcal{X}^A \rightarrow \mathcal{X}^B$)}
    \begin{algorithmic}[1]
        \STATE $\text{\bf{x}}^A_0 \in \mathcal{X}^A, \hat{\text{\bf{x}}}^B_T \sim \mathcal{N}(\text{0},\text{I})$
        \FOR{$t=T,...,t_r+1$}
            \STATE $\boldsymbol\epsilon^A, \boldsymbol\epsilon^B \sim \mathcal{N}(\text{0},\text{I})$
            \STATE $\hat{\text{\bf{x}}}^A_t = \sqrt{\bar{\alpha}_{t^A}}\text{\bf{x}}^A_0+\sqrt{1-\bar{\alpha}_{t^A}}\boldsymbol\epsilon^A$
            \STATE $\hat{\text{\bf{x}}}^B_{t-1} = \frac{1}{\sqrt{1-\alpha_t}}(\hat{\text{\bf{x}}}^B_t-\frac{1-\alpha_t}{\sqrt{1-\bar{\alpha_t}}}\epsilon_{\theta^B}(\hat{\text{\bf{x}}}^B_t, \hat{\text{\bf{x}}}^A_t, t))+\sigma_t \boldsymbol\epsilon^B$
        \ENDFOR
        \FOR{$t=t_r,...,1$}
            \STATE $\boldsymbol\epsilon^A, \boldsymbol\epsilon^B \sim \mathcal{N}(\text{0},\text{I})$ if $t>1$, else $\boldsymbol\epsilon^A, \boldsymbol\epsilon^B=0$
            \STATE $\hat{\text{\bf{x}}}^A_{t-1} = \frac{1}{\sqrt{1-\alpha_t}}(\hat{\text{\bf{x}}}^A_t-\frac{1-\alpha_t}{\sqrt{1-\bar{\alpha_t}}}\epsilon_{\theta^A}(\hat{\text{\bf{x}}}^A_t, \hat{\text{\bf{x}}}^B_t, t))+\sigma_t \boldsymbol\epsilon^A$
            \STATE $\hat{\text{\bf{x}}}^B_{t-1} = \frac{1}{\sqrt{1-\alpha_t}}(\hat{\text{\bf{x}}}^B_t-\frac{1-\alpha_t}{\sqrt{1-\bar{\alpha_t}}}\epsilon_{\theta^B}(\hat{\text{\bf{x}}}^B_t, \hat{\text{\bf{x}}}^A_t, t))+\sigma_t \boldsymbol\epsilon^B$
        \ENDFOR
        \RETURN $\hat{x}^B_0$
    \end{algorithmic}
    \label{alg:translation}
\end{algorithm}
\section{Evaluation}
\label{sec:evaluation}
Our method is evaluated against prior unpaired I2I translation methods~\cite{zhu2017unpaired}~\cite{liu2017unit}~\cite{huang2018multimodal}~\cite{lee2018diverse} on public datasets where ground truth input-output pairs are available~\cite{tylevcek2013spatial}~\cite{zhu2017unpaired}~\cite{hwang2015multispectral}.

\subsection{Baselines}
\begin{figure*}[tb]
    \centering
      \begin{tabular}{c}
  
      \begin{minipage}{0.015\hsize}
          \centering
          \rotatebox[origin=c]{90}{Facade}        
        \end{minipage}
  
        \begin{minipage}{0.13\hsize}
          \centering
            \hspace{0mm} \textnormal{Ground Truth}
            \includegraphics[keepaspectratio, scale=3.1, angle=0]{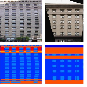}    
        \end{minipage}
  
        \begin{minipage}{0.025\hsize}
          \hspace{2mm}
        \end{minipage}
  
        \begin{minipage}{0.13\hsize}
          \centering
              \textnormal{CycleGAN~\cite{zhu2017unpaired}}
            \includegraphics[keepaspectratio, scale=3.1, angle=0]{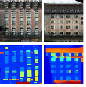}
        \end{minipage}
  
        \begin{minipage}{0.025\hsize}
          \hspace{2mm}
        \end{minipage}
  
        \begin{minipage}{0.13\hsize}
          \centering
              \hspace{0mm} \textnormal{UNIT~\cite{liu2017unit}}
            \includegraphics[keepaspectratio, scale=3.1, angle=0]{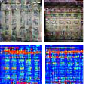}
        \end{minipage}
  
        \begin{minipage}{0.025\hsize}
          \hspace{2mm}
        \end{minipage}
  
        \begin{minipage}{0.13\hsize}
          \centering
              \hspace{0mm} \textnormal{MUNIT~\cite{huang2018multimodal}}
            \includegraphics[keepaspectratio, scale=3.1, angle=0]{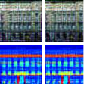}
        \end{minipage}
  
        \begin{minipage}{0.025\hsize}
          \hspace{2mm}
        \end{minipage}
  
        \begin{minipage}{0.13\hsize}
          \centering
              \hspace{0mm} \textnormal{DRIT++~\cite{lee2018diverse}}
            \includegraphics[keepaspectratio, scale=3.1, angle=0]{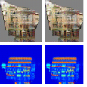}
        \end{minipage}
  
        \begin{minipage}{0.025\hsize}
          \hspace{2mm}
        \end{minipage}
  
        \begin{minipage}{0.13\hsize}
          \centering
              \hspace{0mm} \textnormal{Ours}
            \includegraphics[keepaspectratio, scale=3.1, angle=0]{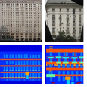}
        \end{minipage} \\
        
  
      \begin{minipage}{0.015\hsize}
          \centering
          \rotatebox[origin=c]{90}{Photos--Maps}        
        \end{minipage}
        
        \begin{minipage}{0.13\hsize}
          \centering
            \includegraphics[keepaspectratio, scale=3.1, angle=0]{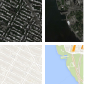}
        \end{minipage}
  
        \begin{minipage}{0.025\hsize}
          \hspace{2mm}
        \end{minipage}
  
        \begin{minipage}{0.13\hsize}
          \centering
            \includegraphics[keepaspectratio, scale=3.1, angle=0]{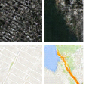}
        \end{minipage}
  
        \begin{minipage}{0.025\hsize}
          \hspace{2mm}
        \end{minipage}
  
        \begin{minipage}{0.13\hsize}
          \centering
            \includegraphics[keepaspectratio, scale=3.1, angle=0]{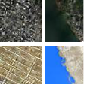}
        \end{minipage}
  
        \begin{minipage}{0.025\hsize}
          \hspace{2mm}
        \end{minipage}
  
        \begin{minipage}{0.13\hsize}
          \centering
            \includegraphics[keepaspectratio, scale=3.1, angle=0]{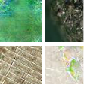}
        \end{minipage}
  
        \begin{minipage}{0.025\hsize}
          \hspace{2mm}
        \end{minipage}
  
        \begin{minipage}{0.13\hsize}
          \centering
            \includegraphics[keepaspectratio, scale=3.1, angle=0]{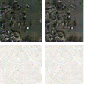}
        \end{minipage}
  
        \begin{minipage}{0.025\hsize}
          \hspace{2mm}
        \end{minipage}
  
        \begin{minipage}{0.13\hsize}
          \centering
            \includegraphics[keepaspectratio, scale=3.1, angle=0]{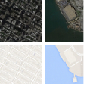}
        \end{minipage} \\

        \begin{minipage}{0.015\hsize}
            \centering
            \rotatebox[origin=c]{90}{Summer--Winter}        
          \end{minipage}
          
          \begin{minipage}{0.13\hsize}
            \centering
              \includegraphics[keepaspectratio, scale=3.1, angle=0]{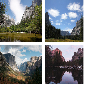}
          \end{minipage}
    
          \begin{minipage}{0.025\hsize}
            \hspace{2mm}
          \end{minipage}
    
          \begin{minipage}{0.13\hsize}
            \centering
              \includegraphics[keepaspectratio, scale=3.1, angle=0]{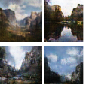}
          \end{minipage}
    
          \begin{minipage}{0.025\hsize}
            \hspace{2mm}
          \end{minipage}
    
          \begin{minipage}{0.13\hsize}
            \centering
              \includegraphics[keepaspectratio, scale=3.1, angle=0]{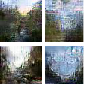}
          \end{minipage}
    
          \begin{minipage}{0.025\hsize}
            \hspace{2mm}
          \end{minipage}
    
          \begin{minipage}{0.13\hsize}
            \centering
              \includegraphics[keepaspectratio, scale=3.1, angle=0]{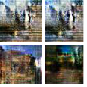}
          \end{minipage}
    
          \begin{minipage}{0.025\hsize}
            \hspace{2mm}
          \end{minipage}
    
          \begin{minipage}{0.13\hsize}
            \centering
              \includegraphics[keepaspectratio, scale=3.1, angle=0]{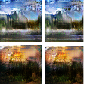}
          \end{minipage}
    
          \begin{minipage}{0.025\hsize}
            \hspace{2mm}
          \end{minipage}
    
          \begin{minipage}{0.13\hsize}
            \centering
              \includegraphics[keepaspectratio, scale=3.1, angle=0]{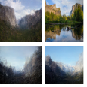}
          \end{minipage} \\

          \begin{minipage}{0.015\hsize}
            \centering
            \rotatebox[origin=c]{90}{RGB--Thermal}        
          \end{minipage}
          
          \begin{minipage}{0.13\hsize}
            \centering
              \includegraphics[keepaspectratio, scale=3.1, angle=0]{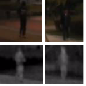}
          \end{minipage}
    
          \begin{minipage}{0.025\hsize}
            \hspace{2mm}
          \end{minipage}
    
          \begin{minipage}{0.13\hsize}
            \centering
              \includegraphics[keepaspectratio, scale=3.1, angle=0]{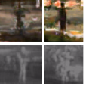}
          \end{minipage}
    
          \begin{minipage}{0.025\hsize}
            \hspace{2mm}
          \end{minipage}
    
          \begin{minipage}{0.13\hsize}
            \centering
              \includegraphics[keepaspectratio, scale=3.1, angle=0]{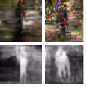}
          \end{minipage}
    
          \begin{minipage}{0.025\hsize}
            \hspace{2mm}
          \end{minipage}
    
          \begin{minipage}{0.13\hsize}
            \centering
              \includegraphics[keepaspectratio, scale=3.1, angle=0]{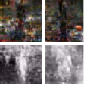}
          \end{minipage}
    
          \begin{minipage}{0.025\hsize}
            \hspace{2mm}
          \end{minipage}
    
          \begin{minipage}{0.13\hsize}
            \centering
              \includegraphics[keepaspectratio, scale=3.1, angle=0]{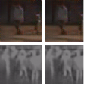}
          \end{minipage}
    
          \begin{minipage}{0.025\hsize}
            \hspace{2mm}
          \end{minipage}
    
          \begin{minipage}{0.13\hsize}
            \centering
              \includegraphics[keepaspectratio, scale=3.1, angle=0]{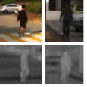}
          \end{minipage}

      \end{tabular}
      \caption{The examples of the output images generated by different image-to-image translation methods.}
      \label{fig:outputs}
  \end{figure*}

The inferred output imagery from our proposed method is compared with that of CycleGAN~\cite{zhu2017unpaired}, UNIT~\cite{liu2017unit}, MUNIT~\cite{huang2018multimodal}, and DRIT++~\cite{lee2018diverse} both quantitatively (Tables~\ref{tab:fid}) and qualitatively (Figures~\ref{fig:outputs}).

\begin{figure}[tb]
  \begin{center}
    \includegraphics[clip,width=8.3cm]{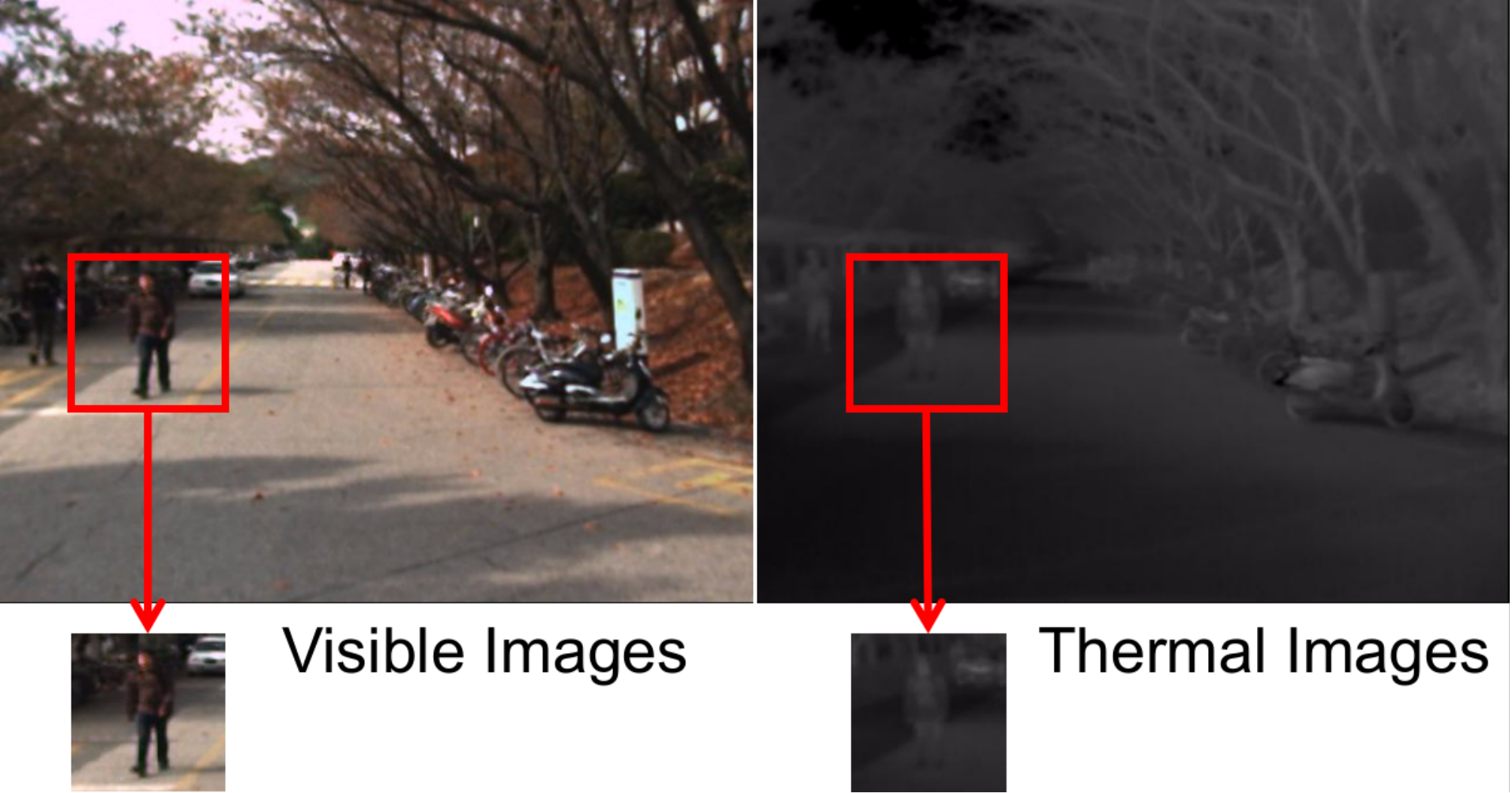}
    \caption{RGB--Thermal dataset cropped from the KAIST Multispectral Pedestrian Dataset\cite{hwang2015multispectral}.}
    \label{fig:kaist}
  \end{center}
\end{figure}

  \begin{figure*}[tb]
  \begin{center}
    \includegraphics[clip,width=17cm]{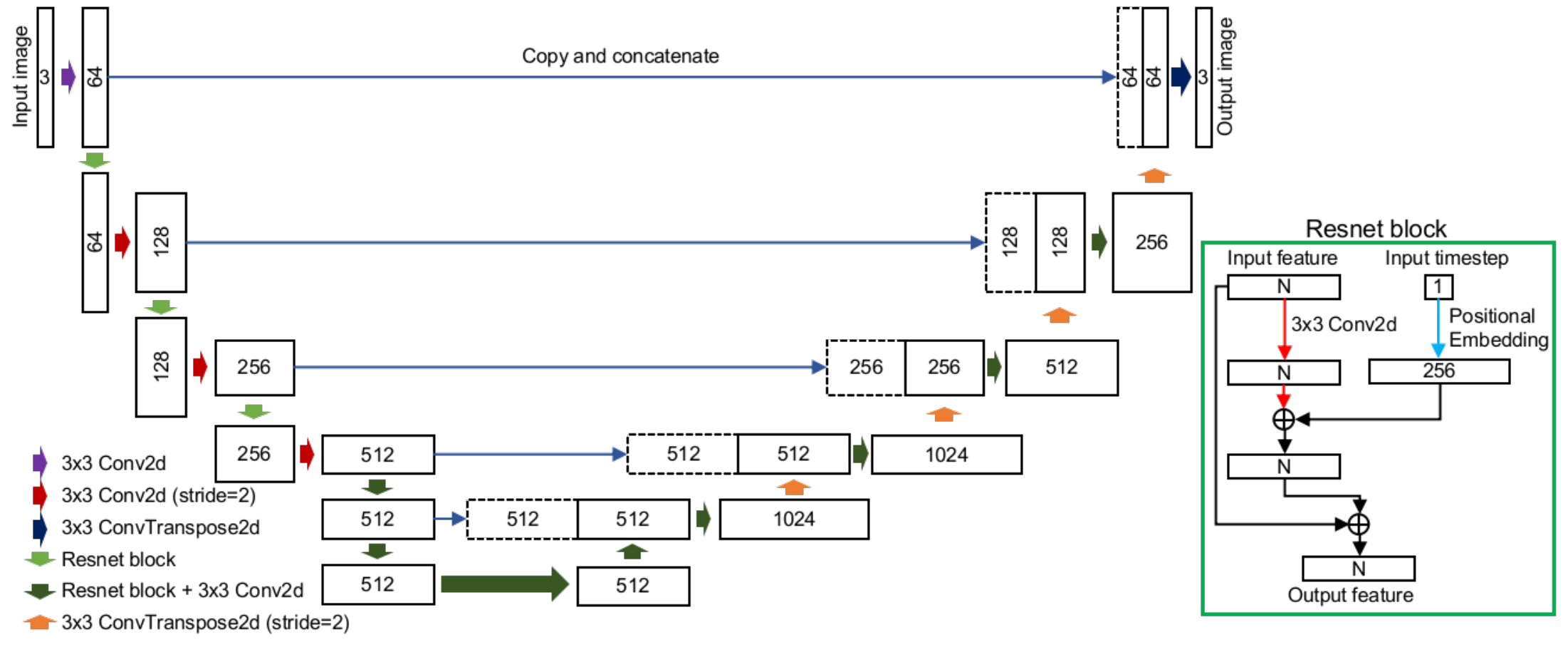}
    \caption{The diagram of our U-net architecture. Each Conv2d or ConvTranspose2d includes BatchNorm2d and ReLU before its input.}
    \label{fig:unet}
  \end{center}
\end{figure*}

  \begin{figure*}[tb]
    \begin{center}
      \includegraphics[clip,width=17.5cm]{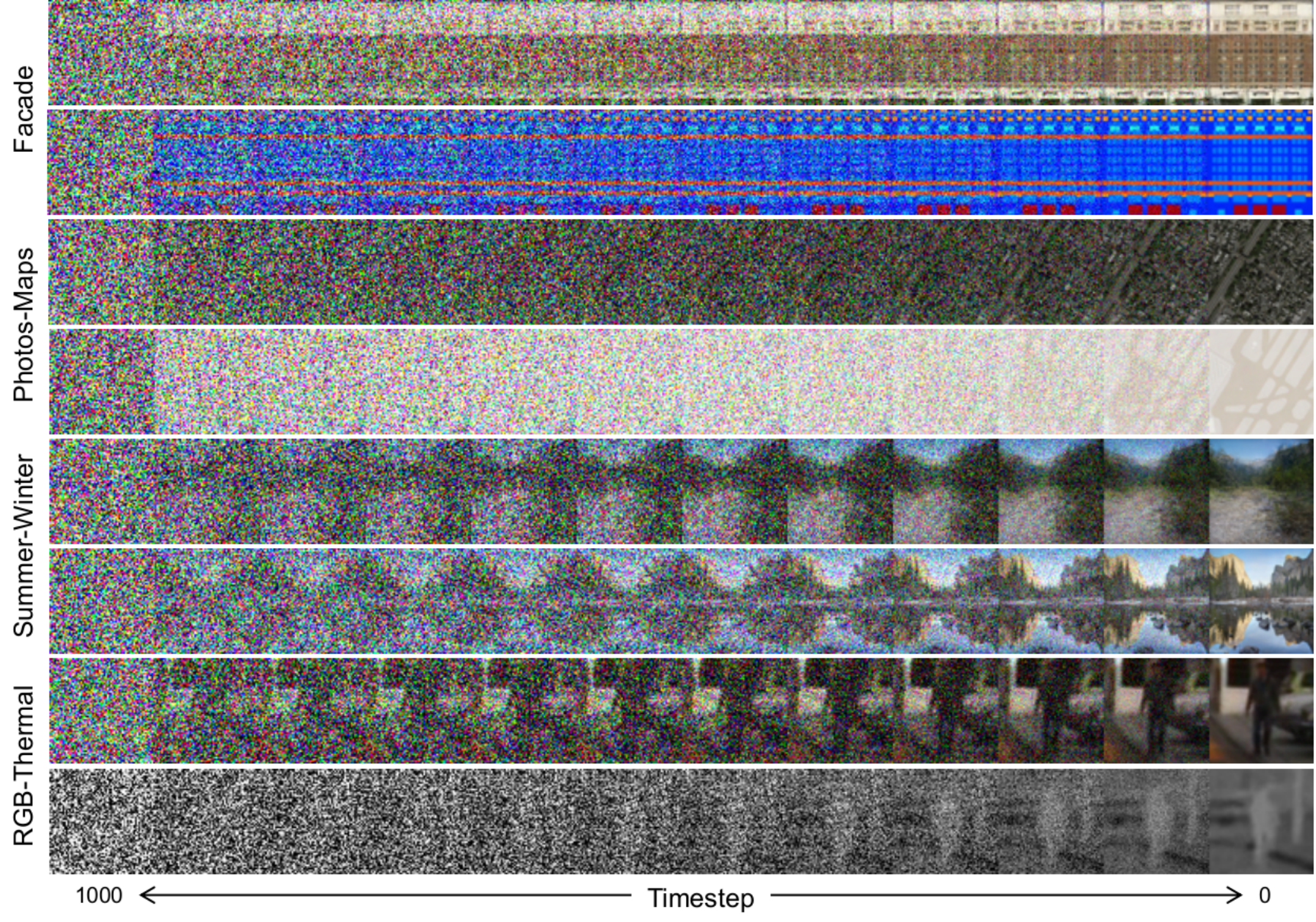}
      \caption{Examples of the progressive image generation via our method.}
      \label{fig:progressive}
    \end{center}
  \end{figure*}

  \subsection{Datasets}
  We prepare the following datasets for the experiment. Every dataset includes two domains (here abbreviated to domain A and B) of images and separated into training and test datasets.
  All images are resized 64 $\times$ 64 pixels in advance.
  \paragraph{Facade}
  The (A) photo and (B) semantic segmentation label images of buildings from the CMP Facades dataset~\cite{tylevcek2013spatial}.
  400 pairs are included for training and 106 pairs for test. 
  \paragraph{Photos--Maps}
  The (A) photo and (B) map images scraped from Google Maps~\cite{zhu2017unpaired}.
  1,096 pairs are included for training and 1,098 pairs for test.
  \paragraph{Summer--Winter}
  The (A) summer and (B) winter Yosemite images downloaded using Flickr API~\cite{zhu2017unpaired}.
  The dataset includes 1,231 summer and 962 winter images for training and 309 summer and 238 winter images for test.
  \paragraph{RGB--Thermal}
  The (A) visible and (B) thermal infrared images of pedestrians from the KAIST Multispectral Pedestrian Dataset~\cite{hwang2015multispectral}.
  This dataset contains aligned visible and thermal images in various regular traffic scenes. Since the images are annotated the region of the pedestrians by bounding boxes, we crop 723 pairs of the pedestrian areas (more than 64 $\times$ 64 pixels size) from the one scene (set00) for training and 425 pairs from another scene (set06) for test (Figure~\ref{fig:kaist}).

  \subsection{I2I translation via UNIT-DDPM}
  \label{sec:i2i_ours}
  The denoising models of our method are implemented using U-Net~\cite{ronneberger2015u} based on PixelCNN~\cite{salimans2017pixelcnn} and Wide ResNet~\cite{zagoruyko2016wide}, Transformer sinusoidal position embedding~\cite{vaswani2017attention} to encode the timestep, whose length is $T=1000$, and $\alpha_t$ is linearly decreased from $\alpha_1=0.9999$ to $\alpha_T=0.98$, as same as original DDPM~\cite{ho2020denoising} but replaced Swith~\cite{ramachandran2017searching} with ReLU~\cite{nair2010rectified}, group normalization~\cite{wu2018group} with batch normalization~\cite{ioffe2015batch}, and removed self-attention block to reduce the computation (Figure~\ref{fig:unet}). The domain translation functions have ResNet~\cite{he2015deep} architecture which has same layer depth of the U-net. In training, the pair of the training sample and the fake another domain sample is concatenated as the input. The model parameters are updated with \mbox{$\lambda_{\text{cyc}} = 10.0$}, batch \mbox{size $B = 16$}, 20,000 epochs via Adam~\cite{kingma2015adam} (initial learning \mbox{rate $\eta= 10^{-5}$}, \mbox{$\beta_1 = 0.5$}, \mbox{$\beta_2 = 0.999$}).

We generate the images in both domains using the trained models and the test samples in each dataset with the \textit{release time} $t_r=1$.

  \subsection{Result}
The output images synthesised by each method are shown in Figure~\ref{fig:outputs} from which it is clearly apparent that our approach qualitatively generates more realistic images than CycleGAN~\cite{zhu2017unpaired}, UNIT~\cite{liu2017unit}, MUNIT~\cite{huang2018multimodal}, and DRIT++~\cite{lee2018diverse}. We also found our method did not suffer from mode collapse at all and the resultant model training was more stable due to not requiring adversarial training. In addition, Figure~\ref{fig:progressive} shows the progressive sampling via our method over the course of the reverse process.
Comparison is conducted via Fréchet Inception Distance (FID)~\cite{heusel2017gans} between the ground truth and output images, and shown in Table~\ref{tab:fid}. Our method outperforms the contemporary approaches of CycleGAN~\cite{zhu2017unpaired}, UNIT~\cite{liu2017unit}, MUNIT~\cite{huang2018multimodal}, and DRIT++~\cite{lee2018diverse} over all of the benchmark datasets Facade, Photos--Maps, Summer--Winter, and RGB--Thermal offering a significant average increase in performance of $\small{\sim} 20\%$ against previous approaches across all such datasets.

\subsection{Ablation Study}
\begin{figure}[tb]
  \begin{center}
    \includegraphics[clip,width=8.5cm]{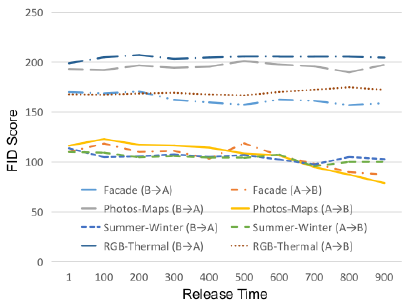}
    \caption{The comparison of FID by the release times.}
    \label{fig:ablation}
  \end{center}
\end{figure}

We analyse the impact of the \textit{release time} against the performance by changing from $t_r=1$ to $900$.
The comparisons of the FID (Figure~\ref{fig:ablation}) show there are no significant change. We can observe slight differences attributable to the \textit{release time} variation, but this is dataset dependent. This result suggests that the tuning of the \textit{release time} hyperparameter is dataset dependent, from which further analysis represents a direction for future work.

\subsection{Limitations}
\begin{figure}[tb]
  \begin{center}
    \includegraphics[clip,width=8.5cm]{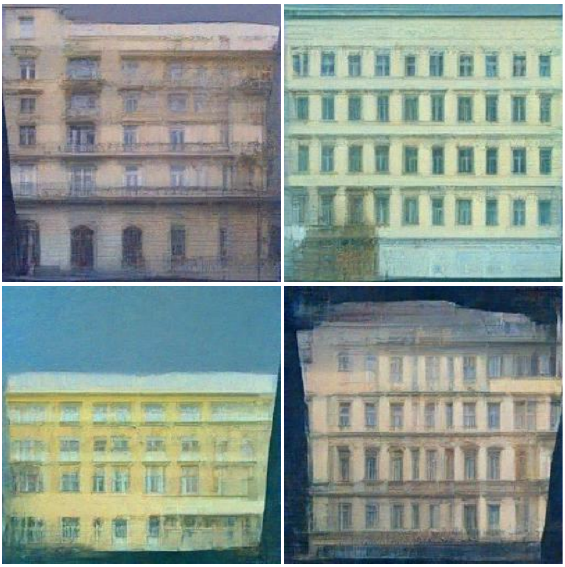}
    \caption{The examples of the output images of 256 $\times$ 256 pixels generated using the model trained by our method (Facade dataset resized to 256 $\times$ 256 pixels).}
    \label{fig:256}
  \end{center}
\end{figure}

We also observe the output images when the input image resolution is increased by 256 $\times$ 256 pixels. The higher resolution models are trained using same network architecture (Figure~\ref{fig:unet}) and learning parameters as Section.~\ref{sec:i2i_ours}. The outputs shown in Figure~\ref{fig:256} are wrongly coloured across the entire pixels. This suggests that the model fails to learn the global information of the images due to the increased complexity of the higher dimensional image space. One of the possible solution is adding more layers along with an attention mechanism into the U-net in the denoising models in order to capture much accurate multi-resolutional structure of images, which will be investigated in future work.

\section{Conclusion}
\label{sec:conclusion}
This paper proposes a novel unpaired I2I translation method that uses DDPM without adversarial training, named UNpaired Image Translation with Denoising Diffusion Probablistic Models (UNIT-DDPM). Our method trains a generative model to infer the joint distribution of images over both domains as a Markov chain by minimising a DSM objective conditioned on the other domain. Subsequently, the domain translation models are simultaneously updated to minimise this DSM objective. After jointly optimising these generative and translation models, we generate target domain images by a denoising MCMC approach, which is conditioned on the input source domain images, based on Langevin dynamics. Our approach provides stable model training for I2I translation and generates high-quality image outputs.
The experimental validation of our approach provides state-of-the-art FID score performance on several public datasets including both colour and multispectral imagery, significantly outperforming contemporary state of the art I2I translation methods.

Although the experiment shows compelling results, the current form of our method is far from uniformly positive in particular in the case of larger resolution. To address this issue, the implementation needs to be modified to model large images much accurately. 

In addition, one remaining drawback of DDPM is the inference time for image generation. However, this can be accelerated by modifying the Markovian process such as denoising diffusion implicit models~\cite{song2020denoising} or reducing the timesteps using a learnable $\Sigma_\theta$~\cite{nichol2021improved}.

Future work will consider modifications to enable shorter sampling times and higher quality image outputs, and the evaluation of the performance when the synthesised images are applied to other downstream computer vision tasks such as object classification.

{\small
\bibliographystyle{ieee_fullname}
\bibliography{references.bib}
}

\end{document}